# Deep Learning and Control Algorithms of Direct Perception for Autonomous Driving

Der-Hau Lee, Kuan-Lin Chen, Kuan-Han Liou, Chang-Lun Liu, Jinn-Liang Liu

*Abstract* — Based on the direct perception paradigm of autonomous driving, we investigate and modify the CNNs (convolutional neural networks) AlexNet and GoogLeNet that map an input image to few perception indicators (heading angle, distances to preceding cars, and distance to road centerline) for estimating driving affordances in highway traffic. We also design a controller with these indicators and the short-range sensor information of TORCS (the open racing car simulator) for driving simulated cars to avoid collisions. We collect a set of images from a TORCS camera in various driving scenarios, train these CNNs using the dataset, test them in unseen traffics, and find that they improve earlier algorithms and controllers in terms of training efficiency and driving stability. Source code and data are available on our website.

## I. Introduction

The direct perception model proposed by Chen et al. [1] maps an input image (high dimensional pixels) from a sensory device of a vehicle to fourteen affordance indicators (a low dimensional representation) by a convolutional neural network (CNN). A controller then drives the vehicle autonomously using these indicators in an end-to-end and real-time manner. This paradigm falls between and displays the merits [1, 2, 3] of the mediated perception [4, 5, 6, 7, 8] and behavior reflex [9, 10, 11, 12, 13] paradigms. We refer to these papers, some recent review articles [14, 15, 16, 17, 18], and many references there for more thorough discussions about these three major paradigms in the state-of-art machine learning (ML) algorithms of autonomous driving.

We shall instead discuss the interplay between CNN and controller and its effects on the overall performance of self-driving cars in training and testing phases, which are not addressed in earlier studies. CNN is a perception mapping from sensory input to affordance output. Controller then maps key affordances to driving actions, namely, to accelerate, brake, or steer [1].

These two mapping algorithms are generally proposed and verified separately since automotive control systems are very complex varying with vehicle types and levels of automation [14, 15, 16, 17, 18]. A great variety of simulators have been developed for simulation testing of autonomous cars in various aspects such as mobility dynamics, path planning, urban traffic, freeway traffic, traffic scene, and safety assessment [17]. However, there are very few [19] open source simulators like TORCS (The Open Racing Car Simulator) [20, 21] and CARLA (Car Learning to Act) [19] that can be used to develop an end-to-end simulating platform with both ML and controller tools for research investigation and verification.

Chen et al. have developed the open source platform named DeepDriving [1] that integrates the CNN AlexNet [22] to TORCS. This platform allows real-time simulation of a CNN pre-trained ego agent (called Host here) driving along with other TORCS agents (called Agents). The main difference between Host and Agents is that they use estimated and true affordance indicators, respectively, to autonomously control their own driving dynamics. It is even more importantly that DeepDriving allows researchers to extend, improve, or verify ML as well as control algorithms in a consistent and unambiguous way.

We propose here fewer affordance indicators than those in [1] and several control algorithms using sensory data to avoid collisions. We show that the extended DeepDriving platform can be used to evaluate different CNNs on equal footing and the modified controller can avoid collisions for Host and Agents in testing phase.

## II. Related Work

Chen et al. [1] proposed 14 indicators (heading angle, 5 distances to preceding cars, 7 distances to lane markings in a three-lane highway, and a Boolean "fast" that is not optimized by the gradient descent method) in two coordinate (in lane and on lane) systems. They only used camera images as inputs to AlexNet [22] modified for autonomous driving (denoted by AlexNet+14 herein). The output estimated indicators from AlexNet+14 are then used in a controller that drives Host in a TORCS traffic with (up to 12) Agents that use true indicators. They also proposed a controller logic for driving Host and Agents. They have collected about 450,000 images from 12 hours of human driving in TORCS video game on 7 different tracks. The maximum speed of Host in their end-to-end simulation is 72 km/h (62 mph).

Based on DeepDriving, Al-Qizwini et al. [2] proposed 5 indicators (heading angle and 4 distances to lane markings), where the 5 distance indicators to preceding cars in [1] are removed and two coordinate systems are reduced to one. In addition to cameras, they used other sensory devices in Host (like lidar and long and short range radars in real cars) to

This work was supported by the Ministry of Science and Technology, Taiwan under Grant MOST107-2115-M-007-017-MY2.

D.-H. Lee is with the Department of Electrophysics, National Chiao Tung University, Taiwan. K.-L. Chen, K.-H. Liou, C.-L. Liu, and J.-L. Liu[*]are with the Institute of Computational and Modeling Science, National Tsing Hau University, Taiwan. [*]Corresponding author email: jlliu@mx.nthu.edu.tw; website: http://www.nhcue.edu.tw/~jinnliu/.

replace these 5 indicators and thus provide the controller more accurate measures of surrounding Agents. They have compared GoogLeNet [23], VGGNet [24], and Clarifai [25] and shown that GoogLeNet performs the best for the root mean squared error (RMSE) of their 5 indicators in training phase. They have also compared GoogLeNet to AlexNet+ with the original 14 and their 5 indicators and shown that RMSEs of these three models are comparable in between 0.01 and 0.02 with GoogLeNet slightly better. The controller has been modified to detect Agents within 60 meters from Host using its sensors and to allow the speed of three Agents larger than that of Host. They collected 510,112 images from 14 hours of a label-collecting agent but did not publish the data and code.

Sauer et al. [3] generalized the direct perception approach to include high-level driving commands such as "turn left at the next intersection" provided by sensor devices in advanced navigation systems [26]. They proposed 6 affordance indicators, namely, heading angle, distance to the vehicle ahead, distance to lane centerline, red light, speed sign, and hazard stop to deal with complex urban environments. The loss function is defined as the sum of the mean absolute error of the first three indicators and the cross entropy of the last three. They have developed a controller that decouples longitudinal control (throttle, brake) using the car-following model in [1] with a proportional–integral–derivative controller and lateral control (steering) using the Stanley controller. The maximum speed of Host in their simulation using CARLA [3] with a front facing camera is 20 km/h in single-lane traffic in driving direction.

## III. CONTROL AND CNN ALGORITHMS

We use a front facing camera and a few short-range radars to design CNN and control algorithms and show that radars can improve driving stability defined by the damage model in TORCS, where the damage number is a measure of an agent colliding with other agents or road obstacles [21].

### A. Affordance Indicators and Control Algorithms

We propose to use three types of indicators, namely, the heading angle of Host (**Angle**), its distance to the road centerline (**toMiddle**), and its distance to the direct preceding car in a certain lane $i$ (**D$i$**). For example, there are 5 indicators, i.e., Angle, toMiddle, D1, D2, and D3 on a three-lane road as shown in Fig. 1.

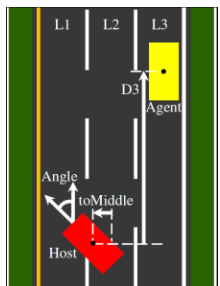

Fig. 1. Five affordance indicators Angle (Host's heading angle), toMiddle (Host's distance to the road centerline), D1 (Host's distance to the direct preceding Agent in Lane 1 (L1)), D2, and D3 on a three-lane highway.

The indicator toMiddle is a critical value for all cars in TORCS to steer and drive on the track [20]. It is used in [3] but not in [1, 2]. The indicators D1, D2 and D3 are different from the five indicators in Figs. 3c and 3e in [1], since we use only one coordinate system as in [2] instead of two.

Controllers are critical in normal and race autonomous driving [1, 2, 3, 20, 21, 27, 28, 29]. We modify TORCS controller, which imitates human drivers in racing with sensor information [27, 28, 29], for highway driving [1, 2] based on these indicators. Our controller is a result of a series of simulation tests targeting at zero damages in TORCS. We investigate the controller in [1] and find collisions taking place due to a lack of using the sensor information of neighboring Agents in Host.

In particular, we find that the following state information of Agents in Host provided by the sensors of TORCS [21] is essential for our learning algorithms to achieve zero damages in our simulation test.

*Agent_State0*: There are no Agents in a range of 60 meters.
*Agent_State1*: A slower Agent is directly in front of Host within the range and with overtaking distance for Host.
*Agent_State2*: A slower Agent is directly in front of Host without overtaking room for Host. This state concerns front-rear collisions.
*Agent_State3*: An Agent is very close to Host in the lateral direction. This state concerns lateral collisions.

Using these four Agent states from TORCS and five indicators from CNN, we design a controller with eight algorithms in Appendix. It can avoid collisions by controlling Host's steering (Algorithms A1, A2, and A3), accelerating when overtaking (Algorithms A4 and A5), and braking when in crowded traffic without room for changing lanes (Algorithms A6 and A7). In these algorithms, we use the sensor information of the speed, location, and yaw angle of neighboring cars. Advanced sensors are vital in autonomous driving and can offer positioning accuracy to a few centimeters [30].

### B. CNN Algorithms

AlexNet+14 in [1] modifies the original AlexNet [22] by switching pooling and local response normalization layers and adding one more fully-connected (FC) layer so that the last four FC layers have 4096, 4096, 256, and 14 units. AlexNet+14 yields estimated (real) values of 14 affordance indicators from an input image. The estimated values are then measured in the mean absolute error (MAE) with the ground truth values provided by TORCS. MAEs are reduced by the stochastic gradient descent optimizer in training phase.

We retain 14 output neurons for 5 indicators in order to compare our CNNs and controller with those in [1], where the absolute error of the indicator toMiddle is weighted 9 times that of the other four. In our experience, toMiddle is more important to learn than the others as its better values yield better Host dynamics in the driving stability of following, curving, and overtaking. The errors of the indicators can be corrected by short-range TORCS sensors to achieve zero damages. The maximum speeds of Agents and Host are 72 and 74 km/h, respectively.

The authors in [23] proposed nine inception modules for GoogLeNet. Each module consists of 1×1, 3×3 and 5×5 convolution kernels and one 1×1 projection layer, which can capture map features at different scales and reduce dimensions and thus remove computational bottlenecks effectively. Moreover, GoogLeNet has a global average pooling after the last convolution layer, which averages out the channel values across the convolutional feature map and hence reduces the total number of parameters drastically. In general, the parameters and memory of a pre-trained GoogLeNet are ~6 million and ~20 MB, respectively, compared to ~60 million and ~200 MB for AlexNet [22]. The authors in [2] used the original GoogLeNet.

We modify GoogLeNet (called GoogLeNet+) by adding one FC layer (having 128 units) in between the sigmoid and FC layers, where three sigmoid layers lead to two auxiliary and one main Euclidean loss layers (see, e.g., Appendex A in [2] for GoogLeNet architecture).

## IV. EXPERIMENTAL SETUP

A key issue in supervised ML is data collection and labeling. For simulation-based autonomous driving research, data comes from a host car by a human driver or a robotic (AI) agent. An AI agent can be thought as a perfect human-like driver [27, 28, 29]. We used an AI agent to collect ~500,000 labeled images on seven different tracks for training as shown in Fig. 6 in [1].

There are three driving scenarios for collecting our data. First, a host agent drives on seven empty tracks in a zigzag manner for training the principal indicators Angle and toMiddle. Second, the host follows closely another very slow and zigzagging AI car in front. Third, the host drives normally on tracks with other AI cars (up to 20). The host drives on each track multiple times to collect data. To obtain different traffic images, AI agents are programmed with various driving behaviors.

We use a different track in testing phase to assess the pre-trained CNN agent with the ground truth data collected by itself. The test data contains about 3000 images. For the assessment, we use MAE of the indicators predicted by CNN algorithms on this static data (sMAE). MAE can also be computed dynamically during driving (dMAE). However, dMAE is larger than sMAE due to asynchronous frequencies between CNN computing (complexity, computer, and speed dependent) and TORCS image generation (computer and speed dependent). CNN and TORCS frequencies are ~15 and >30 Hz, respectively, on our computer.

## V. RESULTS

We train GoogLeNet and GoogLeNet+ using a fine-tuning technique [31] to adapt the self-driving problem. We first train the scratch network by stochastic gradient descent with the batch size bs = 32, the momentum m = 0.9, and the learning rate starting from lr = 0.01 and decreasing by a factor of 0.96 every 32000 iterations. The training process stops after 50k iterations as shown in the inset in Fig. 2. Since the pre-trained network captures general features in its early layers [31], we only fine tune the last FC layers in the re-training process, i.e., all previously trained weights are used as initial guesses except that of the last FC layers set to zero. The loss curve in Fig. 2 shows that error fluctuations (i.e., spiky peaks in the inset) are effectively reduced by fine tuning and errors decrease sharply within 1k iterations. The total number of iterations for GoogLeNet by this two-step training is only 70k to reach sMAE ≈ 0.01 compared to 300k in [2] with comparable errors. Each point in all loss curves is an average of 11 consecutive errors.

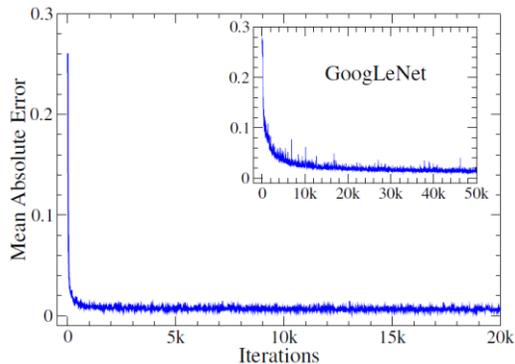

Fig. 2. Mean absolute errors of the five indicators determined by GoogLeNet in two-step training first from scratch (in the inset) with 50k and then from fine-tuning with 20k iterations.

The training losses of GoogLeNet+ and AlexNet+ are shown in Figs. 3 and 4 having sMAE ≈ 0.02 with 120k and 200k iterations, respectively. The hyperparameters of AlexNet+ are the same as those in [1] except m = 0.5.

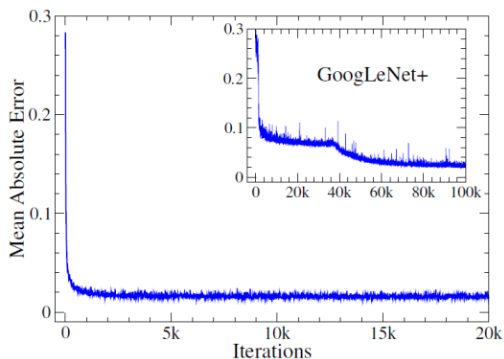

Fig. 3. Mean absolute errors of the five indicators determined by GoogLeNet+ in two-step training first from scratch (in the inset) with 100k and then from fine-tuning with 20k iterations.

GoogLeNet is the best among these three CNNs in training loss. However, GoogLeNet+ yields better Angle and toMiddle in testing phase as shown in Table 1, where Angle is in radians and others are in meters. Angle and toMiddle are principal indicators for Host driving stably around curves, in overtaking, to avoid collissions, and in following as shown in Algorithms A1, A2, A3, and A4, respectively. Larger errors in D1, D2, and D3 are corrected by the sensor information of Agents in Host as shown in Algorithms A1, A2 and A8.

TABLE 1
Static mean absolute errors in testing phase.

| Indicator | AlexNet+ | GoogLeNet | GoogLeNet+ |
|---|---|---|---|
| Angle | 0.034 | 0.041 | 0.029 |
| toMiddle | 0.539 | 0.389 | 0.347 |
| D1 | 6.864 | 5.190 | 6.055 |
| D2 | 7.048 | 3.227 | 3.155 |
| D3 | 8.388 | 5.905 | 5.450 |

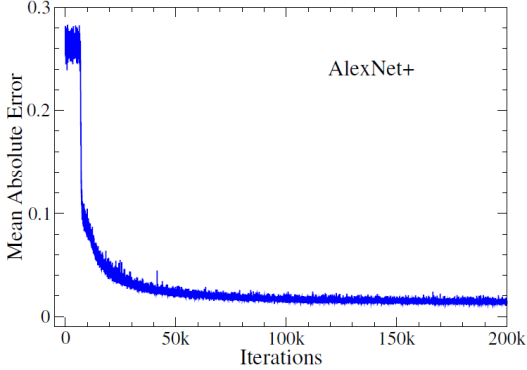

Fig. 4. Mean absolute errors of the five indicators determined by AlexNet+ in one-step training with 200k iterations.

Table 2 shows that the control Algorithms A1 to A8 using these 5 indicators can yield zero collisions for all three CNNs (denoted by CNN5) after one loop on the test track, whereas the original controller in [1] incurs 413 damage points.

TABLE 2
Damage points in testing phase.

| CNN | Damage |
|---|---|
| AlexNet+14 | 413 |
| AlexNet+5, GoogLeNet5, GoogLeNet+5 | 0 |

Finally, we show in Table 3 the dMAEs of 13 indicators in [1] and our 5 indicators obtained by AlexNet+, where distLL = D1, distMM = D2, distRR = D3, and toMiddle is a combined indicator for the two road marking (distL and distR) and seven lane marking (denoted by toMarkX) indicators in [1]. AlexNet+14 is better than AlexNet+5 in dMAE since AlexNet+14 uses two coordinate systems that are however more complicated in designing robust control algorithms to avoid collisions, especially without using sensory information. These difficulties lead to present CNN+5 and control algorithms.

TABLE 3
Dynamic mean absolute errors in testing phase.

| AlexNet+14 | | AlexNet+5 | |
|---|---|---|---|
| Indicator | dMAE | Indicator | dMAE |
| Angle | 0.035 | Angle | 0.043 |
| distLL | 7.970 | D1 | 8.315 |
| distMM | 6.188 | D2 | 9.233 |
| distRR | 8.540 | D3 | 10.198 |
| distL | 2.870 | toMiddle | 0.397 |
| distR | 2.822 | | |
| toMarkL | 0.319 | | |
| toMarkM | 0.374 | | |
| toMarkR | 0.314 | | |
| toMarkLL | 0.291 | | |
| toMarkRR | 0.252 | | |
| toMarkML | 0.257 | | |
| toMarkMR | 0.261 | | |

## VI. CONCLUSION

We have presented five affordance indicators for the direct perception approach to autonomous driving, modified AlexNet and GooLeNet, and proposed several control algorithms that integrate these indicators with the sensors and effectors in TORCS to improve the performance of CNN and TORCS cars in simulated highway traffic. The improvement is based on a quantitative study of the training and testing errors of the modified networks and the damage count of these agent cars using the proposed controller. Our results show that the modified networks can be efficiently trained to infer stable and reliable driving, and that the controller achieves zero collisions for all agent cars in testing phase.

## APPENDIX: CONTROL ALGORITHMS

TORCS provides not only sophisticated physical and 3D graphical engines but also many sensors (*angle*, *speed*, *opponents*, *damage* etc.) and effectors (*steer*, *accel*, *brake* etc.) [27] for model developers to design a variety of normal driving and racing controllers in simulated self-driving traffic [1, 27, 28, 29, 32]. These sensors and effectors are customized to simulate corresponding electronic or mechanical devices in state-of-the-art vehicles [30].

We integrate the indicators *Angle, toMiddle, D1, D2*, and *D3* (whose estimated and true values are used by Host and Agents, respectively), the effectors *steer*, *accel*, and *brake* (whose values are determined by our and TORCS algorithms for Host and Agents, respectively), and the sensor *opponents* (for calculating the *Agent_State0* to *Agent_State3* of Agents) in the following algorithms to design a controller that yields zero *damage* for Host as well as all 20 Agents in testing phase.

For example, we modify the original STEER algorithm of TORCS to Algorithm A1 that returns a value of the effector *steer* using the input values of the five indicators by calling the procedure GETOFFSET in Algorithm A2 that in turn calls Algorithm A8 for the values of *Agent_State0* to *Agent_State3* and determines whether Host should overtake or stay in the current lane and slow down. Algorithm A1 changes the input value of *Angle* (i.e., the controller steers) according to the position of Host, i.e., using the input value of *toMiddle* that determines Host to overtake from left firstly or from right secondly if its value is positive and larger than (the left lane to Host is available for overtaking) or negative and smaller than (the right lane is available) *lane_width*.

---

Algorithm A1: Calculating Steer Value
1: procedure STEER(Angle, toMiddle)
2:   offset=GETOFFSET(D1, D2, D3); //Algorithm A2
3:   Angle−= (toMiddle−offset)/road_width;
     //lane_width=4m, road_width=13m
4:   if (toMiddle>lane_width) then
5:     Angle−= (toMiddle−lane_width)/road_width;
6:   else if (toMiddle<−lane_width) then

```
7:     Angle−=(toMiddle+lane_width)/road_width;
8:     else Angle−=toMiddle/road_width;
9:     steer=Angle/steer_Lock; //steer_Lock=0.366rad
10:    steer=FILTERS(steer); //Algorithm A3
11:    return steer;
12: end procedure
```

Algorithm A2: Calculating Offset for Overtaking
```
1: procedure GETOFFSET(D1,D2,D3)
2:    AGENTSTATE(); //Algorithm A8
3:    if (Agent_State==1) then
4:      if (Agent's toMiddle>1.5) then
5:        offset−=parms; //Agent is near or in left lane
6:      else if (Agent's toMiddle<−1.5) then
7:        offset+=parms; //Agent is near or in right lane
8:      else
9:        if (D2<10) then //middle lane is occupied
10:           if (D1>10) then
11:              offset+=parms;
12:           else if (D1<10 && D3>10) then
13:              offset−=parms;
14:        else offset goes slowly to zero;
15:    else offset goes slowly to zero;
16:    return offset;
17: end procedure
```

Algorithm A3: Steering Filter for Collision Avoidance
```
1: procedure FILTERS(steer)
2:    AGENTSTATE(); //Algorithm A8
3:    if (Agent_State==3) then
4:      if (Agent is near) then
5:        diff_yaw=agent_yaw−host_yaw;
           //agent_yaw given by TORCS
6:        psteer=diff_yaw/steer_Lock;
7:        steer=parm1*steer+parm2*psteer;
8:    return steer;
9: end procedure
```

Algorithm A4: Calculating Acceleration Value
```
1: procedure ACCEL(Angle)
2:  allowed_speed=ALLOWEDSPEED(Angle);
    //Based on estimated road tangent angle=Angle
    //+Host's yaw angle. Speed in m/s.
3:    if (current_speed>allowed_speed) then
4:      accel=allowed_rpm/current_rpm;
5:    else accel=1;
6:    accel=TCS(accel); //AlgorithmA5
7:    return accel;
8: end procedure
```

Algorithm A5: Traction Control System
```
1: procedure TCS(accel)
2:  slip=driven_wheels_speed−current_speed;
3:  if (slip>TCSslip) then
4:    accel−=min(accel,(slip−TCSslip)/TCSrange)
5:  else //TCSslip=2m/s
6:    accel=accel; //TCSrange=10m/s
7:  return accel;
```
8: end procedure

Algorithm A6: Calculating Brake Value
```
1: procedure BRAKE()
2:   if (current_speed>allowed_speed) then
3:     brake=min(1, current_speed −allowed_speed);
4:   else
5:     brake=0;
6:   AGENTSTATE(); //Algorithm A8
7:   if (Agent_State==2) then brake=1;
8:   brake=ABS(brake); //Algorithm A7
9:   return brake;
10: end procedure
```

Algorithm A7: Anti-lock Braking System
```
1: procedure ABS(brake)
2:   if (current_speed>ABSspeed) then
3:     slip=current_speed−avg_4wheels_speed;
4:     if (slip>ABSslip) then
5:       brake−=min(brake,(slip −ABSslip)/ABSrange)
6:   else //ABSspeed=3 m/s
7:     brake= brake; //ABSslip=2 m/s
8:   return brake; //ABSrange=5 m/s
9: end procedure
```

Algorithm A8: Agent State Determination
```
1: procedure  AGENTSTATE()
2:   check D_exact;  //Exact distance between Host and
                     //Agent given by TORCS
3:   Agent_State=0;
4:   if (D_exact<60 && D_exact>−60) then
5:     if (D_exact>4.5) then
6:       Agent_State=1;
7:       if (Host and Agent in same lane) then
8:         if (needed brake distance>D_exact) then
9:           Agent_State=2;
10:    else if (D_exact<4.5 && D_exact>−4.5) then
11:      Agent_State=3;
12:   return Agent_State;
13: end procedure
```